\documentclass{article}

\usepackage[final]{neurips_2020}

\usepackage[utf8]{inputenc} 
\usepackage[T1]{fontenc}    
\usepackage{hyperref}       
\usepackage{url}            
\usepackage{booktabs}       
\usepackage{amsfonts}       
\usepackage{nicefrac}       
\usepackage{microtype}      
\usepackage{amsmath}
\usepackage{caption} 
\usepackage[pdftex]{graphicx}
\usepackage{mathtools}
\usepackage{wrapfig}
\usepackage{subcaption}
\usepackage{natbib}
\usepackage{enumitem}

\DeclareMathOperator*{\argmax}{arg\,max}

\title{Measuring non-trivial compositionality \\ in emergent communication}

\vspace{-30px}
\author{
  Tomasz Korbak\thanks{Currently at Department of Informatics, University of Sussex, United Kingdom.} 
  \\
  Human Interactivity and Language Lab\\
  Faculty of Psychology\\
  University of Warsaw, Poland \\
  \texttt{tomasz.korbak@gmail.com} \\
   \AND
    Julian Zubek \\
  Human Interactivity and Language Lab\\
  Faculty of Psychology\\
  University of Warsaw, Poland \\
  \texttt{j.zubek@uw.edu.pl} \\
     \And
    Joanna Rączaszek-Leonardi \\
  Human Interactivity and Language Lab\\
  Faculty of Psychology\\
  University of Warsaw, Poland \\
  \texttt{raczasze@psych.uw.edu.pl} \\
}
\vspace{-20px}
\begin{document}

\maketitle
\vspace{-20px}

\begin{abstract}
    Compositionality is an important explanatory target in emergent communication and language evolution. The vast majority of computational models of communication account for the emergence of only a very basic form of compositionality: {\it trivial compositionality}. A compositional protocol is trivially compositional if the meaning of a complex signal (e.g. \texttt{blue circle}) boils down to the intersection of meanings of its constituents (e.g. the intersection of the set of blue objects and the set of circles). A protocol is non-trivially compositional (NTC) if the meaning of a complex signal (e.g. \texttt{biggest apple}) is a more complex function of the meanings of their constituents. In this paper, we review several metrics of compositionality used in emergent communication and experimentally show that most of them fail to detect NTC  --- i.e. they treat non-trivial compositionality as a failure of compositionality. The one exception is tree reconstruction error, a metric motivated by formal accounts of compositionality. These results emphasise important limitations of emergent communication research that could hamper progress on modelling the emergence of NTC.
\end{abstract}
\vspace{-5px}
\section{Introduction}

Compositionality is an important explanatory target in emergent communication and language evolution as well as a goal in representation learning and natural language processing \citep{brighton_compositional_2002,lake_building_2016,chaabouni_compositionality_2020}. The vast majority of computational models of communication account for the emergence of only a very basic form of compositionality: trivial \citep{steinert-threlkeld_towards_2020} or na\"ive compositionality \citep{kharitonov_emergent_2020}. Natural languages are non-trivially compositional (NTC) as they include phenomena like quantifiers, negation, word order and context-dependence. A communication protocol is NTC if for a certain complex signal (e.g. \texttt{biggest apple}) its meaning is {\it not} just the intersection of the meanings of its constituents but some more complex function of the constituents. Despite being a necessary milestone towards accounting for the evolution of language, NTC has received little attention from the machine learning and evolutionary linguistics communities. We conjecture that this state of affairs is partly due to an inability to quantitatively measure progress toward NTC in computational settings.

In this study, we review seven metrics of compositionality and experimentally show that most fail to detect NTC --- i.e. they treat non-trivial compositionality as a failure of compositionality. We do observe, however, that properly parametrised tree reconstruction error (TRE) \citep{andreas_measuring_2019} --- a metric directly motivated by a formal account of compositionality \citep{montague_universal_1970} --- detects NTC to a significant degree. 

To summarise, the contributions of this paper are (i) providing a common framework for comparing different approaches to measuring compositionality in emergent communication, (ii) experimentally showing that most metrics used in machine learning and evolutionary linguistics fail to detect NTC, and (iii) demonstrating how to parametrise TRE to be able to detect NTC. The anonymised code for all experiments and reusable implementations of all metrics are publicly available from \url{https://github.com/tomekkorbak/measuring-non-trivial-compositionality}.
\vspace{-5px}
\section{Background}

\begin{wrapfigure}{r}{0.5\textwidth}
\centering
\includegraphics[width=1\linewidth]{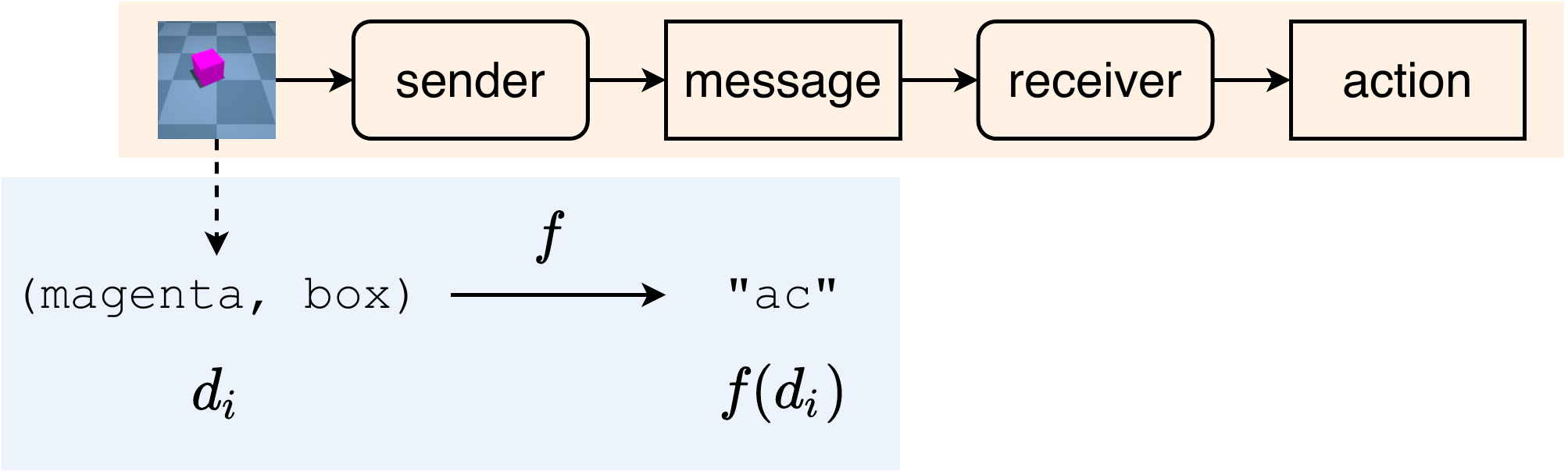}
\caption{Most research on emergent communication assumes the setup displayed on the orange plate: a sender observes an RGB image $x$, sends a message to the receiver and the receiver acts based on the message. To measure compositionality, we can simplify this setup (blue plate) and consider only a derivation $d_i$ describing a compositional structure of a situation and a function $f$ mapping each derivation to a message. A derivation $d_i$ mirrors RGB image $x$ while the representation $f(d_i)$ is identical to the message sent by the sender upon observing $x$.}
\label{diagram1}
\vspace{-20px}
\end{wrapfigure}

Most research on emergent communication has focused on a Lewis signalling game of the following form: a sender sends a message to a receiver upon observing a situation and the receiver acts based upon the message \citep{lewis_convention:_1969,skyrms_signals:_2010}. If the incentives of the sender and the receiver are aligned, they will agree on a communication protocol. To simplify the problem of measuring compositionality, let us assume that the situations observed by the sender are governed by an underlying compositional structure (known to us but hidden from the sender) and let us focus on the communication protocol itself, understood as a mapping from those hidden compositional structures (henceforth called derivations) to messages (henceforth called representations) as shown in Figure \ref{diagram1}.

\paragraph{Derivations} A \emph{derivation} $d \in \mathcal{D}$ can be thought of as a tree representing a situation. Derivations are defined recursively such that
if $d_i$ and $d_j$ are derivations, then $(d_i, d_j)$ is a derivation, where $(\cdot, \cdot)$ is a \emph{derivation composition} function. A primitive derivation $d_0$ is called a \emph{concept}. For instance, a blue circle corresponds to a derivation $(\texttt{blue}, \texttt{circle})$ built out of two concepts: $\texttt{blue}$ and $\texttt{circle}$. In an emergent communication setting, that derivation can be the structure underlying an RGB image observed by the sender.

\paragraph{Representations} Let us have a set of \emph{representations} $\theta \in \Theta$. A representation can be thought of as a description of a derivation. Representations can be composed together, e.g. $\theta_i = \theta_j \circ \theta_k$, where $\theta_j, \theta_k \in \Theta$ and $\circ$ is a \emph{representation composition} function. A primitive representation $\theta_0$ is called a \emph{symbol}. Finally, a communication protocol is a function $f: \mathcal{D} \to \Theta$ mapping derivations to representations.

In this paper, we will consider three perspectives of what kind of object a representation $\theta \in \Theta$ is:
\vspace{-7px}
\begin{enumerate}[leftmargin=*]
    \item From a communication perspective, $\Theta$ is the set of messages understood as strings over an alphabet $\Sigma$, i.e. $\Theta = \Sigma^*$. Then, $\circ$ corresponds to string concatenation. For instance, $f((\texttt{blue}, \texttt{circle})) = \texttt{ab}$.
    \item From a semantic perspective, $\Theta$ is the set of meanings associated with derivations. We will assume meanings to be sets of objects such as circles or boxes. Then, $\circ$ corresponds to a function over sets (e.g. set intersection). For instance, $f((\texttt{blue}, \texttt{circle})) = \{ x : \texttt{blue}(x) \land \texttt{circle}(x) \}$.
    \item From a geometric perspective, $\Theta$ is a vector space. Then, $\circ$ corresponds to vector addition. For instance, $f((\texttt{blue}, \texttt{circle})) \in \mathbb{R}^N$, where $N = \text{dim} \ \Theta$.
\end{enumerate}

While we ultimately care about the communication perspective, defining the distinction between trivial compositionality and NTC requires the semantic perspective and measuring compositionality in terms of TRE requires approximating the semantic perspective in terms of the geometric perspective.

\paragraph{Compositionality} Intuitively, a communication protocol embodied by $f$ is \emph{compositional} if the space of representations is homeomorphic to the space of derivations. More formally, $f$ is compositional if the following holds:
\begin{equation}
    \label{comp}
    f((d_i, d_j)) = f(d_i) \circ f(d_j)
\end{equation}

In other words, the composition function over representations $\circ$ mirrors the composition function over derivations $(\cdot, \cdot)$: for each derivation $d$ obtained by applying operator $(\cdot, \cdot)$ its image $f(d)$ can be obtained by applying a corresponding operator $\circ$.

This mathematical model of compositionality, originally constructed by \cite{montague_universal_1970} using universal algebra, is the dominant approach in formal semantics (see \citep{janssen_compositionality_2010} for a review). In the context of emergent communication, this model was recently explicitly assumed by \cite{andreas_measuring_2019} and \cite{steinert-threlkeld_towards_2020}.

\paragraph{Trivial and non-trivial compositionality} Let us take the semantic perspective and assume $\Theta$ to be a set of sets of objects. Then, a communication protocol is \emph{trivially} compositional (TC) if the representation composition function $\circ$ is set intersection. Alternatively, $f$ is NTC if $\circ$ is a more complex function over sets of objects \citep{steinert-threlkeld_towards_2020}.

Most signalling games studied in machine learning and evolutionary linguistics are confined to TC communication protocols. For instance, a communication protocol defined over objects with shapes and colours would probably be TC, with the meaning of a message $\texttt{blue circle}$ being the intersection of the set of circle with the set of blue objects  \citep{mordatch_emergence_2017,kottur_natural_2017,korbak_template_transfer_2019}. On the other hand, a great deal of natural language semantics is NTC. For instance, the meaning of the phrase $\texttt{good cook}$ is \emph{not} the intersection of the set of cooks with the set of good people. Rather, the adjective $\texttt{good}$ is highly contextual and complements the meaning of the noun  $\texttt{cook}$ differently than it complements the meaning of, for example, the noun $\texttt{climber}$.

\paragraph{Tree reconstruction error} 

TRE is a metric of compositionality proposed by \cite{andreas_measuring_2019} and directly motivated by Montague's account of compositionality embodied in \eqref{comp}. First, assume there is a distance function over representations $\delta: \Theta \times \Theta \to \mathbb{R}_+$. Then we can define a \emph{compositional approximation}  $\hat{f}_\phi$ of $f$ with parameters $\phi$ as follows:
\begin{equation}
    \label{tre}
    \begin{cases*}
        \hat{f}_\phi(d_i) = \phi_i & for $d_i \in \mathcal{D}_0$, \\
        \hat{f}_\phi((d_j, d_k)) = \hat{f}_\phi(d_j) \circ \hat{f}_\phi(d_k) & for all other $d$.
    \end{cases*}
\end{equation}

In other words, $\hat{f}_\phi$ assigns each $d \in \mathcal{D}_0$ an embedding vector and composes these vectors using $\circ$ for complex derivations. $\circ$ can be a non-parametric vector operation, e.g. addition, or a parametric transformation, e.g. a linear transformation. The parameters $\phi$ (embedding vectors for concepts and possible parameters of $\circ$) are optimised so we have
\begin{equation}
    \phi^* = \argmax_\phi \mathbb{E}_{d \sim \mathcal{D}} \ \delta(\hat{f}_\phi(d), f(d)). 
\end{equation}
The irreducible distance $\delta(\hat{f}_{\phi^*}(d), f(d))$ given the optimal parameters $\phi^*$ is the TRE.

Unlike in a signalling game, while optimising $\hat{f}_\phi$ we do have explicit access to the underlying derivation $d \in \mathcal{D}$. Therefore what TRE measures is how well a given communication protocol can be reconstructed while respecting the compositional structure of $d$. A compositional protocol satisfying \eqref{comp} will by definition respect $d$, and hence can be reconstructed perfectly. 
\vspace{-5px}

\section{Experiments}

In our experiments, we consider a well-studied signalling game in which the sender observes objects endowed with two discernible features: shape and colour. The corresponding derivations $d \in \mathcal{D}$ are ordered tuples of two kinds of concepts: shapes and colours, e.g. $(\texttt{blue}, \texttt{circle})$. The set of primitive derivations $\mathcal{D}_0 = \{\texttt{blue}, \texttt{red}, \dots\} \cup \{ \texttt{square}, \texttt{triangle}, \dots \}$ consists of 25 colours and 25 shapes. We take the set of representations to be a set of strings of length $L$ over a finite alphabet, i.e. $\Theta = \Sigma^L$, where $\Sigma = \{a, b, \dots \}$.

We consider nine pre-defined communication protocols suitable for solving the signalling game defined above: one TC, six NTC (entangled, diagonal, negation, rotated, context-sensitive) and two non-compositional baselines (random and holistic). We designed these protocols as minimal models of NTC phenomena found in natural languages and formal languages: negation (e.g. $\texttt{not circle}$), conversational context (e.g. requiring only shape to be communicated), word order ($\texttt{ab}$ is different from $\texttt{ba}$) and entanglement in the representation learning sense \citep{kharitonov_emergent_2020}. These \emph{probing} protocols are thus aligned with linguistic intuitions (and with existing literature, whenever possible) about what constitutes (trivial or non-trivial) compositionality. For a detailed description of all communication protocols considered in this experiment, see appendix \ref{proto}.

We then consider seven metrics of compositionality used by the machine learning community and report how they score the protocols. The metrics considered are TRE, conflict count, topographic similarity, BOW disentanglement, generalisation, positional disentanglement and context independence. For TRE, we implemented $\circ$, the composition functions for $N$-dimensional vector representations of derivations, as linear transformation --- i.e. $ \circ: s_1 \circ s_2 \mapsto A s_1 + B s_2$,
 where $s_1, s_2 \in \Theta$ are vector-encoded symbols and $A, B \in \mathbb{R}^{N \times N}$ are learnable parameters. We describe experiments with other implementations of $\circ$ in appendix \ref{tre-circ}. For a detailed descriptions of used metrics, see appendix \ref{metrics}. 

\begin{figure}[h]
\centering
\includegraphics[width=\linewidth]{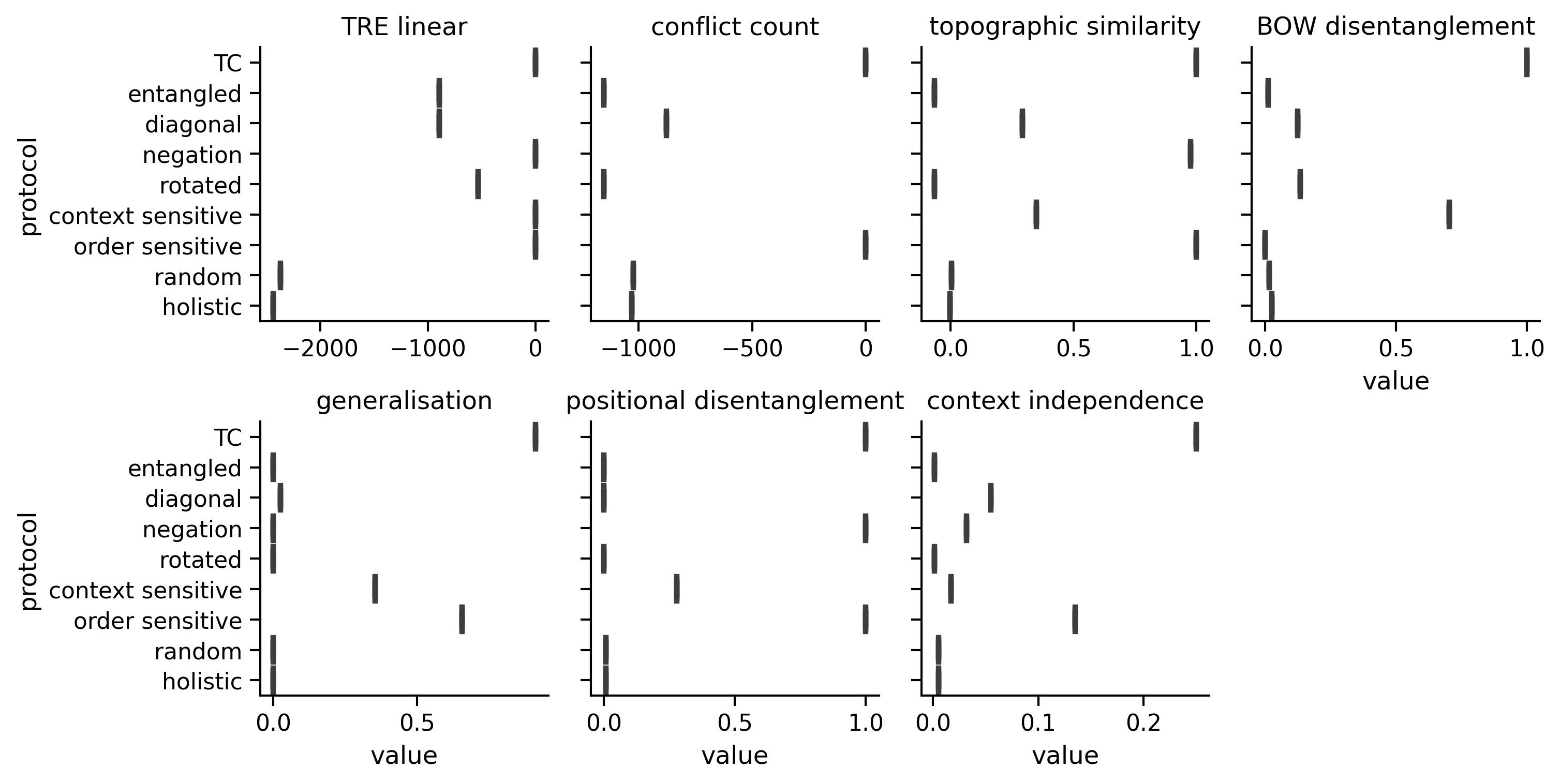}
\caption{Scores assigned by various compositionality metrics to various protocols. Each subplot corresponds to a different metric, with protocols on the Y-axis and scores on the X-axis. Note that the X-axes does not share a common scale and we report \emph{negative} TRE and \emph{negative} conflict count. Thus, for each metric, a higher value means greater compositionality. Conflict count is undefined for negation and context-sensitive protocols. We observed negligible variance across across five random seeds.}
\label{plot1}
\end{figure}

The results of the evaluation are displayed in Figure \ref{plot1}. We can observe that while (almost) all protocols assign high compositionality scores to the TC protocol and low compositionality scores to non-compositional (holistic and random) protocols, most also assign low scores to NTC protocols. Generalisation, somewhat in line with recent results \citep{chaabouni_compositionality_2020}, is low for some NTC protocols (negation, context-sensitive, entangled, diagonal, rotated). Assuming that receiver's generalisation requires both productivity of a communication protocol used by the sender and capacity of the receiver, low generalisation for NTC protocols may be explained by insufficient capacity of the receiver. This, in turn, suggests that generalising to NTC requires higher capacity and/or stronger inductive biases than generalising to TC. Context independence, topographical similarity, positional disentanglement, BOW disentanglement and conflict count can pick up only the simplest forms of NTC such as negation and and order-sensitivity. TRE is the only metric that assigns high scores to all NTC protocols.

\clearpage

\vspace{-15px}
\section{Discussion}

We conjecture that the reason most metrics fail to capture NTC is because they were designed under the assumption of TC as the canonical form of compositionality. This assumption, however, seems to be guided by a simplified signalling game setup rather than formal accounts of compositionality \citep{janssen_compositionality_2010} or corpus data. That design choice may, in turn, stem from a more general problem of designing meaningful metrics for emergent communication \citep{lowe_pitfalls_2019} and translating theoretical accounts in linguistics into quantitative measures. To illustrate one such translation problem, if we do not restrict the composition function $\circ$ to be of a particular class, any language may be considered compositional (see appendix \ref{tre-circ}).

Despite these difficulties, NTC is ubiquitous in natural languages. Phenomena such as function words and dependency relations \citep{rizzi_functional_2016} demonstrate that primitive concepts cannot be treated as completely orthogonal \citep{murphy_comprehending_1988} and natural languages use more than one form of symbol composition \citep{gardenfors_language_1995}. Our aim in this paper was to introduce NTC as an explanatory target for emergent communication and to demonstrate how to measure it in terms of TRE. We hope these contributions will guide future work accounting for the emergence of NTC and closing the gap between emergent and human communication.


%
%

\section*{Broader impact}

The field of emergent communication constitutes basic research and is focused on a theoretical problem: the emergence of language. However, the problem of learning compositional representations and understanding compositional language has broader implications for natural language processing and representation learning. Concrete problems that can be posed as emergent communication include image captioning \citep{kottur_natural_2017}) and unsupervised machine translation \citep{lee_emergent_2017} (both of which can be considered visually grounded communication) as well as explainability \citep{andreas_translating_2017}. Research on learning compositional representations also informs the development of natural language processing technologies such as semantic parsing and machine reasoning \citep{hudson_compositional_2018}. These systems are vulnerable to bias, permit malicious use and can give rise to unintended adverse effects. On the other hand, the conjectured interpretability and robustness of compositional representations could improve the transparency and fairness of machine learning systems that utilise such representations, as well as advance progress on conversational systems that empower disadvantaged individuals.

\section*{Acknowledgments}

Tomasz Korbak, Julian Zubek and Joanna Rączaszek-Leonardi were funded by a National Science Centre (Poland) grant OPUS 2018/29/B/HS1/00884. The authors are grateful to Krzysztof Główka, Łukasz Kuciński and Paweł Kołodziej for their helpful feedback.

\bibliography{ntc_refs}

\clearpage
\appendix

\section{Communication protocols used in the experiments}
\label{proto}

Recall that given an observation based on derivation $d \in \mathcal{D}$ the sender sends a message $m$ composed of $L$ symbols $(s_1, s_2, \dots, s_L) \in \Sigma^L$:

\begin{equation}
    f(d) = (s_1, s_2).
\end{equation}

Unless specified otherwise, $L = 2$, $d = (d_1, d_2)$, where $d_1, d_2 \in \mathcal{D}_0$ are primitive observations for shape and colour. Then, each message has a form $(s_1, s_2)$. 

The communication protocols $f$ are defined as follows.

\paragraph{Trivially compositional (TC) protocol} The TC protocol was constructed by assuming a fixed one-to-one mapping between concepts and symbols, e.g. $\texttt{blue}$ with $\texttt{a}$ and $\texttt{circle}$ with $\texttt{b}$ and generating a message for each shape--colour pair by concatenating associated symbols, e.g. $f((\texttt{blue}, \texttt{circle})) = \texttt{ab}$.

\paragraph{Holistic protocol} In animal communication research, a \emph{holistic} communication system is one in which messages (e.g. $(s_1, s_2)$) only have meaning as a whole and their parts ($s_1$ and $s_2$) are meaningless without context. Hence, we construct a holistic communication protocol by uniformly sampling a pair of symbols $(s_1, s_2)$ (without replacement) from $\Sigma^2$ for each $d \in \mathcal{D}$.

\paragraph{Random protocol} We construct the random protocol by sampling each symbol separately, with replacement, i.e. $s_1 \sim p(\Sigma), s_2 \sim p(\Sigma)$ for each  each $d \in \mathcal{D}$.

\paragraph{Entangled NTC protocol} We use an example provided by \cite{kharitonov_emergent_2020}, a compositional protocol that refers to complex properties of the objects constructed as combinations of basic concepts.

Let us also assume that both concepts and symbols are represented by non-negative integers from finite fields closed under modular addition and subtraction, e.g. $\mathcal{D}_0 = \{0, 1, 2, \dots, 10\}, \Sigma = \{0, 1, 2, \dots, 10\}$. Then we have:
\begin{equation}
    s_1 = (d_1 - d_2) \mod 10,
\end{equation}
\begin{equation}
    s_2 = (d_1 + d_2) \mod 10.
\end{equation}

The non-triviality of $f$ here stems from the fact that it entangles shapes and colours, so that both $s_1$ and $s_2$ depend on both the shape and the colour. \cite{kharitonov_emergent_2020} consider this protocol to be a counter-example to \emph{na\"ive compositionality}, which is essentially what we mean by NTC.

\paragraph{Rotated NTC protocol} The rotated protocol is similar to the entangled protocol.
It is obtained by encoding concepts numerically, rotating the coordinate system by 45 degrees and
then mapping obtained values to symbols in the order they appear on the newly obtained axes.
Under the same assumptions as the entangled protocol above, the concept--symbol relationship is defined
as:
\begin{equation}
    s_1 = d_1 - d_2 + 10
\end{equation}
\begin{equation}
    s_2 = d_1 + d_2
\end{equation}

\paragraph{Order-sensitive NTC protocol} This protocol is constructed analogously to the TC protocol with one difference: each symbol is used to communicate both a colour and a shape. For instance, $\texttt{d}$ means $\texttt{blue}$ when on the first position in the message and $\texttt{square}$ when on the second position. This protocol is NTC because it constrains $\circ$ to be non-commutative.

\paragraph{Context-sensitive NTC protocol} This protocol is based on the game described by \cite{barrett_hierarchical_2018}. We modify our setup so the derivations $d \in \mathcal{D}$ are nested tuples $(\texttt{context}, (\texttt{colour}, \texttt{shape}))$. $\texttt{context}$ can be $\texttt{colour}$, $\texttt{shape}$ or $\texttt{both}$. This corresponds to a signalling game when the sender is also provided with information which concept (shape and/or colour) must be communicated to the receiver and disincentivised from communicating too much. This protocol is NTC because message length is a function of context.

\paragraph{Negation NTC protocol} The negation protocol is based on the intuition that negation constitutes a minimal instance of NTC in natural language semantics. To illustrate this, we modify our setup by assuming that there are just two shapes, $\texttt{circle}$ and $\texttt{box}$, and the agent only has a word $\texttt{x}$ to refer to the $\texttt{box}$. However, it also uses a symbol $\texttt{!}$ as a negation and can refer to circles as `not boxes', i.e. $\texttt{!x}$. The part of the message communicating the shape is then trivially composed, with the second part communicating colour. The non-triviality of this protocol lies in the fact that the meaning of $\texttt{!x}$ is a nontrivial function of the meaning $\texttt{!}$ and $\texttt{x}$. The semantics of this protocol cannot be formulated in terms of set intersection, because there is no set that could constitute the meaning of the negation token $\texttt{!}$.

\paragraph{Diagonal NTC protocol} This protocol reflects a language with two-word utterances, where one word represents intensity and the second certain property or axis of variability (examples from natural language: `low  brightness', `high contrast', `medium volume'). In this example we refer to a complex property being a combination of basic concepts.

Let us assume both concepts and symbols are represented by non-negative integers: $d_1 \in \{0, 1, \ldots, n_1\}$, $d_2 \in \{0, 1, \ldots, n_2\}$, $s_1 \in \{0, 1, \ldots, n_1 + n_2\}$, $s_2 \in \{0, 1, \ldots, n_1\}$. An object embodying concepts $d_1$, $d_2$ is represented with a pair of symbols $(s_1, s_2)$:

\begin{equation}
    s_1 = d_1 + d_2
\end{equation}
\begin{equation}
    s_2 =
\begin{cases}
d_2 \;\; \mathrm{if} \; d_1 + d_2 \leq n_1\\
n_1 - d_1 \;\; \mathrm{otherwise}
\end{cases}
\end{equation}

Concrete examples of communication protocols described above are provided in Tables \ref{proto1} and \ref{proto2}.

\begin{table}[p]
    \centering
\begin{tabular}{llllllll}
\toprule
derivation&holistic&TC&random&entangled&order sensitive&diagonal&rotated\\
\midrule
$(\texttt{blue},\texttt{square})$&\texttt{ia}&\texttt{hf}&\texttt{fj}&\text{ii}&\texttt{dd}&\texttt{dd}&\texttt{di}\\
$(\texttt{blue},\texttt{circle})$&\texttt{hj}&\texttt{hi}&\texttt{eg}&\text{jd}&\texttt{da}&\texttt{jj}&\texttt{ef}\\
$(\texttt{blue},\texttt{ellipse})$&\texttt{ea}&\texttt{hd}&\texttt{gd}&\text{af}&\texttt{dc}&\texttt{hh}&\texttt{ch}\\
$(\texttt{blue},\texttt{triangle})$&\texttt{bd}&\texttt{ha}&\texttt{ic}&\text{eg}&\texttt{de}&\texttt{bb}&\texttt{hc}\\
$(\texttt{blue},\texttt{rectangle})$&\texttt{fc}&\texttt{hc}&\texttt{if}&\text{cb}&\texttt{db}&\texttt{aa}&\texttt{fe}\\
$(\texttt{green},\texttt{square})$&\texttt{gc}&\texttt{jf}&\texttt{bh}&\text{dd}&\texttt{cd}&\texttt{jd}&\texttt{bf}\\
$(\texttt{green},\texttt{circle})$&\texttt{hg}&\texttt{ji}&\texttt{hf}&\text{if}&\texttt{ca}&\texttt{hj}&\texttt{dh}\\
$(\texttt{green},\texttt{ellipse})$&\texttt{ga}&\texttt{jd}&\texttt{bf}&\text{jg}&\texttt{cc}&\texttt{bh}&\texttt{ec}\\
$(\texttt{green},\texttt{triangle})$&\texttt{gi}&\texttt{ja}&\texttt{hi}&\text{ab}&\texttt{ce}&\texttt{ab}&\texttt{ce}\\
$(\texttt{green},\texttt{rectangle})$&\texttt{ba}&\texttt{jc}&\texttt{dh}&\text{eh}&\texttt{cb}&\texttt{gb}&\texttt{hd}\\
$(\texttt{gold},\texttt{square})$&\texttt{fa}&\texttt{ef}&\texttt{fb}&\text{ff}&\texttt{ed}&\texttt{hd}&\texttt{jh}\\
$(\texttt{gold},\texttt{circle})$&\texttt{ef}&\texttt{ei}&\texttt{ff}&\text{dg}&\texttt{ea}&\texttt{bj}&\texttt{bc}\\
$(\texttt{gold},\texttt{ellipse})$&\texttt{gf}&\texttt{ed}&\texttt{aj}&\text{ib}&\texttt{ec}&\texttt{ah}&\texttt{de}\\
$(\texttt{gold},\texttt{triangle})$&\texttt{fg}&\texttt{ea}&\texttt{ae}&\text{jh}&\texttt{ee}&\texttt{gh}&\texttt{ed}\\
$(\texttt{gold},\texttt{rectangle})$&\texttt{jf}&\texttt{ec}&\texttt{ji}&\text{ac}&\texttt{eb}&\texttt{ch}&\texttt{cb}\\
$(\texttt{yellow},\texttt{square})$&\texttt{gg}&\texttt{gf}&\texttt{fh}&\text{gg}&\texttt{ad}&\texttt{bd}&\texttt{ac}\\
$(\texttt{yellow},\texttt{circle})$&\texttt{ha}&\texttt{gi}&\texttt{aa}&\text{fb}&\texttt{aa}&\texttt{aj}&\texttt{je}\\
$(\texttt{yellow},\texttt{ellipse})$&\texttt{fb}&\texttt{gd}&\texttt{ib}&\text{dh}&\texttt{ac}&\texttt{gj}&\texttt{bd}\\
$(\texttt{yellow},\texttt{triangle})$&\texttt{ig}&\texttt{ga}&\texttt{ia}&\text{ic}&\texttt{ae}&\texttt{cj}&\texttt{db}\\
$(\texttt{yellow},\texttt{rectangle})$&\texttt{ii}&\texttt{gc}&\texttt{gg}&\text{je}&\texttt{ab}&\texttt{ij}&\texttt{ej}\\
$(\texttt{red},\texttt{square})$&\texttt{db}&\texttt{bf}&\texttt{bc}&\text{bb}&\texttt{bd}&\texttt{ad}&\texttt{ge}\\
$(\texttt{red},\texttt{circle})$&\texttt{bf}&\texttt{bi}&\texttt{jd}&\text{gh}&\texttt{ba}&\texttt{gd}&\texttt{ad}\\
$(\texttt{red},\texttt{ellipse})$&\texttt{hh}&\texttt{bd}&\texttt{ch}&\text{fc}&\texttt{bc}&\texttt{cd}&\texttt{jb}\\
$(\texttt{red},\texttt{triangle})$&\texttt{ce}&\texttt{ba}&\texttt{fa}&\text{de}&\texttt{be}&\texttt{id}&\texttt{bj}\\
$(\texttt{red},\texttt{rectangle})$&\texttt{jc}&\texttt{bc}&\texttt{fb}&\text{ia}&\texttt{bb}&\texttt{ed}&\texttt{da}\\
\bottomrule
\end{tabular}
      \vspace{10px}
    \caption{Examples of holistic, TC, random, entangled, order sensitive, diagonal and rotated protocols for derivations with five colours and five shapes.}
      \label{proto1}
\end{table}

\begin{table}[h!]
    \centering
\begin{subtable}[t]{0.48\textwidth}
\begin{tabular}{ll}
\toprule
derivation&message\\
\midrule
$(\texttt{colour},(\texttt{blue},\texttt{square}))$&\texttt{e}\\
$(\texttt{shape},(\texttt{blue},\texttt{square}))$&\texttt{c}\\
$(\texttt{both},(\texttt{blue},\texttt{square}))$&\texttt{ec}\\
$(\texttt{colour},(\texttt{blue},\texttt{circle}))$&\texttt{e}\\
$(\texttt{shape},(\texttt{blue},\texttt{circle}))$&\texttt{h}\\
$(\texttt{both},(\texttt{blue},\texttt{circle}))$&\texttt{eh}\\
$(\texttt{colour},(\texttt{blue},\texttt{ellipse}))$&\texttt{e}\\
$(\texttt{shape},(\texttt{blue},\texttt{ellipse}))$&\texttt{a}\\
$(\texttt{both},(\texttt{blue},\texttt{ellipse}))$&\texttt{ea}\\
$(\texttt{colour},(\texttt{blue},\texttt{triangle}))$&\texttt{e}\\
$(\texttt{shape},(\texttt{blue},\texttt{triangle}))$&\texttt{g}\\
$(\texttt{both},(\texttt{blue},\texttt{triangle}))$&\texttt{eg}\\
$(\texttt{colour},(\texttt{blue},\texttt{rectangle}))$&\texttt{e}\\
$(\texttt{shape},(\texttt{blue},\texttt{rectangle}))$&\texttt{i}\\
$(\texttt{both},(\texttt{blue},\texttt{rectangle}))$&\texttt{ei}\\
$(\texttt{colour},(\texttt{green},\texttt{square}))$&\texttt{d}\\
\bottomrule
\end{tabular}
\caption{Context-sensititive protocol}
\end{subtable}
\begin{subtable}[t]{0.48\textwidth}
\begin{tabular}{ll}
\toprule
derivation&message\\
\midrule
$(\texttt{blue},\texttt{circle})$&$\texttt{a!x}$\\
$(\texttt{blue},\texttt{box})$&$\texttt{ax}$\\
$(\texttt{red},\texttt{circle})$&$\texttt{b!x}$\\
$(\texttt{red},\texttt{box})$&$\texttt{bx}$\\
$(\texttt{green},\texttt{circle})$&$\texttt{c!x}$\\
$(\texttt{green},\texttt{box})$&$\texttt{cx}$\\
$(\texttt{yellow},\texttt{circle})$&$\texttt{d!x}$\\
$(\texttt{yellow},\texttt{box})$&$\texttt{dx}$\\
$(\texttt{gold},\texttt{circle})$&$\texttt{e!x}$\\
$(\texttt{gold},\texttt{box})$&$\texttt{ex}$\\
$(\texttt{orange},\texttt{circle})$&$\texttt{f!x}$\\
$(\texttt{orange},\texttt{box})$&$\texttt{fx}$\\
$(\texttt{white},\texttt{circle})$&$\texttt{g!x}$\\
$(\texttt{white},\texttt{box})$&$\texttt{gx}$\\
$(\texttt{black},\texttt{circle})$&$\texttt{h!x}$\\
$(\texttt{black},\texttt{box})$&$\texttt{hx}$\\
$(\texttt{pink},\texttt{circle})$&$\texttt{i!x}$\\
$(\texttt{pink},\texttt{box})$&$\texttt{ix}$\\
$(\texttt{silver},\texttt{circle})$&$\texttt{j!x}$\\
$(\texttt{silver},\texttt{box})$&$\texttt{jx}$\\
$(\texttt{bronze},\texttt{circle})$&$\texttt{k!x}$\\
$(\texttt{bronze},\texttt{box})$&$\texttt{kx}$\\
\bottomrule
\end{tabular}
\caption{Negation protocol}
\end{subtable}
   \vspace{10px}
    \caption{Examples of context-sensitive and negation protocols}
   \label{proto2}
\end{table}

\section{Compositionality metrics}
\label{metrics}

\paragraph{Generalisation} Compositionality is widely considered to be the feature of language and thought that explains the generalisation capabilities of humans \citep{chomsky_syntactic_1957,lake_building_2016}. While recent research in emergent communication shows that the relationship between compositionality and generalisation is  nuanced \citep{chaabouni_compositionality_2020,kharitonov_emergent_2020} and in some signalling games compositionality is not necessary for generalisation, generalisation to novel situations remains an intuitive hallmark of compositionality. 

Here we measure the test set accuracy of a receiver trained to predict the ground-truth derivations $d$ based on messages send by a fixed sender $f(d)$. More concretely, we implement the receiver as a neural networks that first embeds each symbol of a message $f(d)$ into a 50 embedding vector, feeds each of these embedding to a single-layer LSTM \citep{hochreiter_long_1997} and then feeds the last hidden state vector of the LSTM into a two-layer feed-forward neural network. The output of the network is a tuple of categorical distributions over all concepts in the derivation. The loss function consists is a sum of cross-entropy errors for for all concepts. The neural network is implemented in PyTorch \citep{paszke_automatic_2017} using EGG \citep{kharitonov_egg:_2019}. We train it using Adam \citep{kingma_adam:_2014} with learning rate $10^{-2}$ and batch size 1. We use $\mathcal{L}_2$ regularisation with coefficient $\lambda = 10^{-6}$ and initialise the embedding vectors by sampling from $\mathcal{N}(0, 1)$. 

For the purpose of our experiments, we split the set of derivations $\mathcal{D}$ into $\mathcal{D}_{\text{train}}$ (80\% of the derivations) and $\mathcal{D}_{\text{test}}$ (20\% of the derivations). We train the receiver on $\mathcal{D}_{\text{train}}$ for 100 epochs or until it achieves training set accuracy 1. We then measure the accuracy of the receiver on $\mathcal{D}_{\text{test}}$. The reported accuracies are averaged across five random seeds.

\paragraph{Positional disentanglement} \cite{chaabouni_compositionality_2020} introduced positional disentanglement as an adaptation of similar metrics developed in the representation learning community \citep{chen_isolating_2019}. It is also related to context independence and residual entropy introduced by \cite{resnick_capacity_2020}. Let $s_j$ denote the $j$-th symbol of a message $f(d)$, and $c_1^j$ the concept with the highest mutual information with $s_j$, and $c_2^j$ with the second highest mutual information:
\begin{equation}
    c_1^j = \argmax_c \mathcal{I}(s_j; c),
\end{equation}
\begin{equation}
    c_2^j = \argmax_{c \neq c_1^j} \mathcal{I}(s_j; c),
\end{equation}
where $\mathcal{I}(\cdot; \cdot)$ is mutual information and $c \in \mathcal{D}_0$. Then, positional disentanglement $\text{posdis}$ is defined as
\begin{equation}
    \text{posdis} = \frac{1}{L} \sum_{j=1}^L \frac{\mathcal{I}(s_j; c^j_1) - \mathcal{I}(s_j; c^j_2)}{\mathcal{H}(s_j)}
\end{equation}
where $L$ is the maximum message length and $\mathcal{H}(s_j)$ is entropy over the distribution of symbols at $j$-th place in messages $f(d)$ for each $d \in \mathcal{D}$. We ignore positions with zero entropy.

\paragraph{Bag-of-words disentanglement} Note that positional disentanglement assumes that compositionality involves fixed order (e.g. the meaning of symbol $\texttt{a}$ at first place is different from the meaning of symbol $\texttt{a}$ at second place in the message).  Bag-of-words disentanglement relaxes this assumption by only considering symbols counts: $n_j$ is the number of occurrences in $j$-th symbol in a message. Then, bag-of-words disentanglement, $\text{bosdis}$, is defined as
\begin{equation}
    \text{bosdis} = \frac{1}{|\Sigma|} \sum_{j=1}^{|\Sigma|} \frac{\mathcal{I}(n_j; c^j_1) - \mathcal{I}(n_j; c^j_2)}{\mathcal{H}(n_j)},
\end{equation}
where $|\Sigma|$ is the number of symbols available in the protocol.

\paragraph{Tree reconstruction error}

We define $\hat{f}_\phi$ to be a neural network so we can optimise its parameters $\phi$ via gradient descent over $\nabla_\phi \delta(f(d), \hat{f}_\phi(d))$. More concretely, to generate a reconstruction $\hat{f}_\phi(d)$ of a derivation $d$, we follow \eqref{tre} and first embed each concept $d \in \mathcal{D}_0$ forming $d$ into an $N$-dimensional embedding vector, where $N = |\Sigma| L$ (with $L$ the maximum message length, fixed in advance). Then, we encode the entire $d$ into an $N$-dimensional embedding vector by recursively applying $\circ$ in a bottom-up manner. The ground truth message corresponding to derivation $d$ is encoded as $L$ one-hot vectors. We then define $\delta(f(d), \hat{f}_\phi(d))$ to be a sum of $L$ cross-entropy errors between $i$-th segment of the reconstruction $\hat{f}_\phi(d))$ and $i$-th one-hot-encoded symbol in the ground truth message $f(d)$. The neural network was implemented in PyTorch \citep{paszke_automatic_2017}. We train it for 1000 epochs using Adam \citep{kingma_adam:_2014} with learning rate $10^{-1}$. We use $\mathcal{L}_2$ regularisation with coefficient $\lambda = 10^{-5}$ and initialise the embedding vectors by sampling from $\mathcal{N}(0, 1)$.

\paragraph{Context independence} Context independence  \citep{bogin_emergence_2018} measures the alignment between symbols forming a message and concepts forming a derivations. Let us denote the set of concepts by $\mathcal{D}_0$ and the set of symbols by $\Sigma$. By $p(s|c)$, we mean the probability that $f$ maps a derivation containing concept $c \in \mathcal{D}_0$ to a message containing symbol $s \in \Sigma$. We define the inverse probability $p(c|s)$ similarly. Finally, we define $s^k := \argmax_s p(c \vert s)$; $s^c$ is the symbol most often sent in presence of a concept $c$. Then, context independence metric is $\mathbb{E} ( p(v^k \vert k) \cdot p(k \vert v^k))$; the expectation is taken with respect to the joint uniform distribution $p(\mathcal{D}_0, \Sigma)$. 

For instance, when the derivation consists of a shape and a colour, our experiments, context independence measures the consistency of associating symbols with shapes irrespective of colour and vice versa. Note that context independence effectively punishes the agents for using synonyms, i.e. associating multiple symbols with a single concept \citep{lowe_pitfalls_2019}.  

\paragraph{Topographical similarity} Topographical similarity \citep{brighton_understanding_2006,lazaridou_emergence_2018} is a measure of structural similarity between messages and derivations. Let us define $L_\mathcal{D} : \mathcal{D} \times  \mathcal{D} \to \mathbb{R}_+$ to be a distance over derivations and $L_{m} : \Sigma^* \times  \Sigma^* \to \mathbb{R}_+$ to be a distance over messages. Topographical similarity is the Spearman $\rho$ correlation of $L_t$ and $L_m$ measured over a joint uniform distribution $p(\mathcal{D}_0, \Sigma)$. Topographical similarity mirrors the approach known as \emph{representation similarity analysis} in systems neuroscience \citep{kriegeskorte_representational_2008} where it is used to quantify structural similarity between a stimulus and neural activity evoked by the stimulus

We choose $L_{m}$ to be the \cite{levenshtein_binary_1966} distance and treat derivations as ordered pairs of concepts so we can choose $L_\mathcal{D}$ to be the Hamming distance.

\paragraph{Conflict count} Conflict count was introduced by \cite{kucinski_emergence_2020}. It assumes that the number of concepts in a derivation $K$ is equal to message length $L$ and that there is a one-to-one mapping between each concept $d \in \mathcal{D}_0$ and symbol $s \in \Sigma^L$. It then counts how often this mapping is violated.

Let us denote each permutation mapping the position of a symbol to the position of concepts as $\phi$, where $\phi:\{1, \ldots, L\}\mapsto \{1, \ldots, K\}$. Then, let us denote the principal meaning of a symbol $s\in \Sigma^L$ at position $j$ as $\text{m}(s, j):=\argmax_{d}\text{count}(s, j, d)$, where

\begin{equation}
\text{count}(s, j, d) = \sum_{d' \in \mathcal D} \mathbf{1}(
f(d)_j = s, d'_{\phi(j)} = d
)
\end{equation}
Here $\mathbf{1}$ denotes the indicator function and $d_j$ the $j$-th concept in a derivation $d$. Then, conflict count is
\begin{equation}
\min_\phi \sum_{s, j} \text{score}(s, j; \phi),
\end{equation}
where $\text{score}(s,j;\phi) = \sum_{d\neq \text{m}(s,j)} \mathtt{count}(s,j, d)$.

Because conflict count assumes the number of concepts in a derivation $K$ to be equal to message length $L$, it is undefined for two protocols violating this assumption: negation and context sensitive.

\section{Effect of composition function $\circ$ in TRE}
\label{tre-circ}

\begin{figure}[h]
\centering
\includegraphics[width=0.8\linewidth]{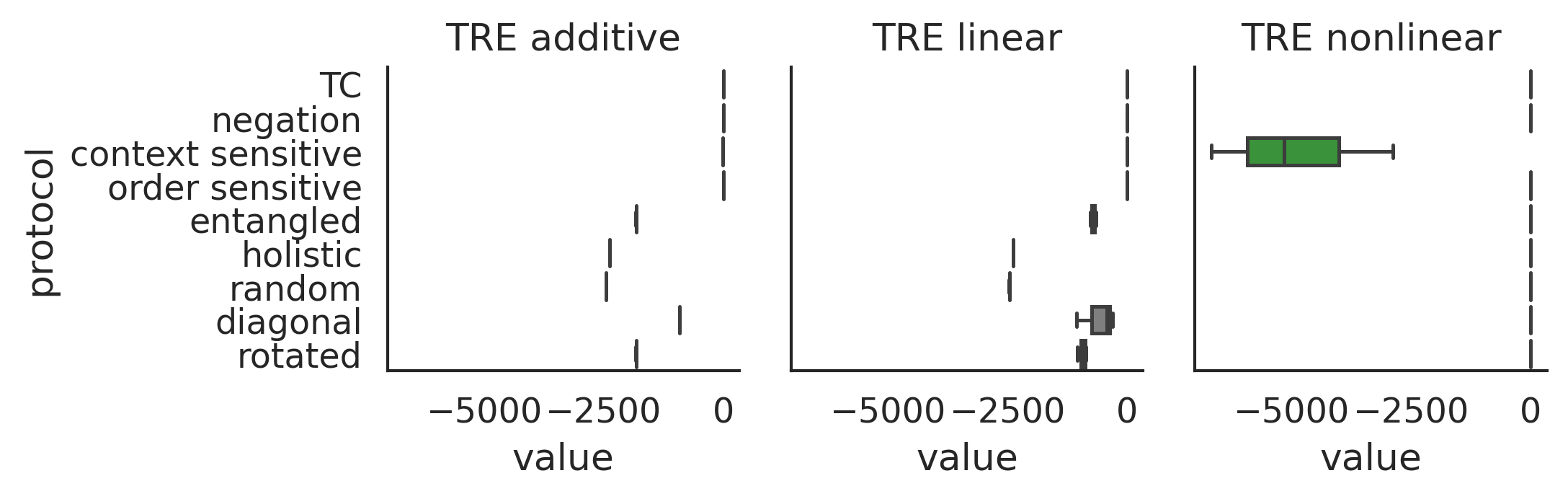}
\caption{Scores assigned by various compositionality metrics to various protocols. Each subplot corresponds to a different metric, with protocols on the Y-axis and scores on the X-axis. Again, the X-axes do not share a common scale and we report \emph{negative} TRE so a higher value means greater compositionality. The width of the box for each score is its confidence interval describing the standard deviation across five random seeds.}
\label{plot2}
\end{figure}

In this additional experiment, we analyse the effect of various implementations of  $\circ$ (the composition function for $N$-dimensional vector representations of derivations $d \in \mathcal{D}$) on TRE scores across protocols. We consider three implementations of $\circ$:
\begin{enumerate}
    \item Additive composition,  where $\circ$ is vector addition: 
    \begin{equation}
        \circ: s_1 \circ s_2 \mapsto s_1 + s_2.
    \end{equation}
    \item Linear composition, where $\circ$ is a linear transformation: 
    \begin{equation}
        \circ: s_1 \circ s_2 \mapsto A s_1 + B s_2,
    \end{equation}
 where $A, B \in \mathbb{R}^{N \times N}$ are learnable parameters.
    \item Non-linear composition, where $\circ$ is a two-layer feedforward neural network: 
    \begin{equation}
        \circ: s_1 \circ s_2 \mapsto W_2\tanh(W_{11}s_1 + W_{12}s_2 + b_1) + b_2.
    \end{equation}
    Here $W_2 \in \mathbb{R}^{N \times H}$, $W_{11}, W_{12} \in \mathbb{R}^{H \times N}$, $b_1 \in \mathbb{R}^H$,  $b_2 \in \mathbb{R}^N$ and $H$ denotes the size of the hidden layer. We choose $H = 50$.
\end{enumerate}
The results of the experiments are presented in Figure \ref{plot2}. While additive and linear composition perform similarly, the model capacity of non-linear composition is probably too strong for the task, resulting in severe overfitting (e.g. low TRE even for random and holisitc protocols) and a false negative for the context-sensitive protocol. The presented results were stable across hyperparameters of TRE (e.g. learning rate, weight decay coefficient, number of epochs).

\end{document}